# The Principles of Human-like Conscious Machine

Fangfang Li
Liff1229@hotmail.com
Mind Simulation Lab

Xiaojie Zhang
zhangxj966@gmail.com
Chongqing Medical and Pharmaceutical College

## Abstract

Determining whether another system—biological or artificial—possesses phenomenal consciousness has long been a central challenge in consciousness studies. This attribution problem has become especially pressing with the rise of large language models and other advanced AI systems, where debates about "AI consciousness" implicitly rely on some criterion for deciding whether a given system is conscious. In this paper, we propose a substrate-independent, logically rigorous, and counterfeit-resistant sufficiency criterion for phenomenal consciousness. We argue that any machine satisfying this criterion should be regarded as conscious with at least the same level of confidence with which we attribute consciousness to other humans. Building on this criterion, we develop a formal framework and specify a set of operational principles that guide the design of systems capable of meeting the sufficiency condition. We further argue that machines engineered according to this framework can, in principle, realize phenomenal consciousness. As an initial validation, we show that humans themselves can be viewed as machines that satisfy this framework and its principles. If correct, this proposal carries significant implications for philosophy, cognitive science, and artificial intelligence. It offers an explanation for why certain qualia—such as the experience of red—are in principle irreducible to physical description, while simultaneously providing a general reinterpretation of human information processing. Moreover, it suggests a path toward a new paradigm of AI beyond current statistics-based approaches, potentially guiding the construction of genuinely human-like AI.

# 1. Introduction

The study of consciousness has long been regarded as one of the most profound and elusive challenges in philosophy and neuroscience. Understanding the nature of consciousness is crucial, as it not only delves into the essence of human experience but also has profound implications for fields ranging from artificial intelligence to ethics (Nagel, 1974; Chalmers, 1996).

In his influential work, David Chalmers distinguishes between what he calls the "easy problem" and the "hard problem" of consciousness. The easy problem refers to the question of how the brain processes stimuli and produces behavior, a challenge that, while complex, is amenable to scientific investigation. In contrast, the hard problem concerns the subjective experience of consciousness itself—why and how certain brain processes give rise to the rich inner world of phenomenal consciousness or qualia, a question that remains largely unexplained and deeply mysterious (Chalmers, 1995).

In his 1995 paper, Chalmers initially identified verbal report as a part of the "easy problem" of consciousness. Verbal reports refer to the ability of individuals to describe their conscious experiences in language, an issue that, while important, Chalmers argued could be addressed through standard scientific methods of investigation (Chalmers, 1995). However, in his 2018 work on the meta-problem of consciousness, Chalmers expanded his view by introducing a new category of "easy problems" that are intimately connected to the hard problem. This new class involves the verbal report of consciousness itself—asking individuals to explain the nature of consciousness or to provide a report on the subjective experience of being conscious. Chalmers suggested that tackling these kinds of questions could, in fact, contribute to the resolution of the hard problem by offering deeper insights into the nature of subjective experience and its neural correlates (Chalmers, 2018).

In this paper, we, first, build upon Chalmers' framework and propose a method, in an ideal scenario, for assessing whether a subject possesses phenomenal consciousness through verbal report. The underlying logic of this method is based on the premise that subjective experience, or qualia, is inherently a private and internal phenomenon—it is something that only the individual subject can directly access. Therefore, subjective experience can be viewed as a form of "hidden information". Thus, we can use the method of determining whether a person knows a specific hidden piece of information to assess whether a machine possesses phenomenal consciousness.

After proposing this method, we begin with constructing a machine that meets these criteria. Specifically, we take inspiration from Kant's philosophical arguments:

> "... we have no concepts of the understanding and hence no elements for the cognition of things except insofar as an intuition can be given corresponding to these concepts, consequently that we can have cognition of no object as a thing in itself, but only insofar as it is an object of sensible intuition, i.e. as an appearance; " (Kant, 1998).

and

> "The same function that gives unity to the different representations in a judgment also gives unity to the mere synthesis of different representations in an intuition, which, expressed generally, is called the pure concept of the understanding" (Kant, 1998).

These arguments can be summarized as follows: 1) Humans cannot directly perceive the true essence of the world; they can only experience its appearances as shaped by their sensory organs; 2)Humans possess innate cognitive structures that organize these sensory appearances into a unified and coherent experience, effectively constructing their perception of reality.

These two insights are broadly consistent with mainstream cognitive science: the human body forms perception by organizing signals from the external world through the perceptual system in an innately biased and experience-shaped way.

Motivated by these insights, we consider how a machine would process information if it were to face the same situation that Kant describes for humans—namely, the inability to directly access the essence of the external world and the necessity of constructing an understanding based on appearances. To address this question, we propose four principles for processing information.

Based on these principles, we argue that what is commonly referred to as qualia is, in fact, the machine's internal representation of objects in the form of signal groups. To support this claim, we demonstrate that these internally represented signal groups exhibit the key properties of qualia—namely, ineffability, physical irreducibility, intentionality, and unity.

Through this approach, we demonstrate that a machine built according to these four information-processing principles can exhibit the key properties of consciousness without receiving any external information about consciousness itself. This means the machine meets the criteria we initially proposed for determining whether a machine possesses consciousness. As a result, we conclude that this machine has phenomenal consciousness.

In the following section, we argue that these four information-processing principles are not arbitrary inventions; rather, they reflect the fundamental principles by which humans process information. To support this claim, we systematically identify corresponding concepts in cognitive science and compare them to our proposed principles.

Specifically, we examine the similarities and differences between the concepts emerging from our principles and those in mainstream cognitive science. This includes comparative discussions on feature binding, attention, and memory indexing. Additionally, we provide empirical evidence to support the implications of these differences.

Through this approach, we demonstrate that it is reasonable to assert that human information processing aligns with the four principles we have proposed.

In summary, the main contributions of this paper are as follows:

1) We propose a method for determining whether a machine possesses phenomenal consciousness.

2) We present the principles of information processing that enable a machine to have phenomenal consciousness.

3) We demonstrate why these principles align with the internal mechanisms of the human mind.

## 2. Discriminative and Explanatory Theories of Consciousness

The scientific study of consciousness can be divided into two complementary theoretical enterprises: discriminative theories and explanatory theories.

Discriminative theories aim to provide criteria for determining whether a given system possesses phenomenal consciousness (PC). They address the attribution problem, or the other-mind problem: given an arbitrary biological or artificial system, how do we decide whether it is conscious? Ideally, a discriminative theory specifies either sufficient or necessary conditions for consciousness or better, an operational criterion that is generalizable across substrates, e.g. Integrated Information Theory (IIT) (Tononi, 2004; Oizumi, Albantakis & Tononi, 2014; Albantakis & et al. 2023,)

Explanatory theories, by contrast, seek to answer why or how certain physical or computational mechanisms give rise to PC. Representative examples in cognitive science include the Global Workspace Theory (GWT) (Baars, 1997, 2007; Baars, Geld & Kozma, 2001; Dehaene, 2014; Dehaene&Changeux, 2011), Higher-Order Thought theories (Rosenthal, 1997, Lau & Rosenthal, 2011; Brown, Lau & LeDoux, 2019;), Local Recurrent Processing theory(Lamme, 2006, 2010, 2020), and Attention Schema Theory (AST) (Graziano & Kastner 2011; Graziano 2013, 2019;). They all attempt to explain consciousness in terms of certain neural mechanisms.

Methodologically, explanatory theories can only be empirically meaningful if they are coupled with discriminative criteria. Without a way to decide which systems are conscious, empirical data cannot confirm or disconfirm a proposed mechanism. The development of robust discriminative theories is therefore not optional but a prerequisite for scientific progress on consciousness.

## 2.1 The Philosophical Landscape and the Need for Criteria

Robert Kuhn's "A landscape of consciousness: Toward a taxonomy of explanations and implications" (2024) offers a helpful taxonomy, spanning eliminative materialism, epiphenomenalism, functionalism, emergentism, biological naturalism, property dualism, and panpsychism. These frameworks differ in their metaphysical commitments, but as soon as they make a differentiating claim that some entities are conscious while others are not, they face the same challenge: how to justify this distinction empirically.

Illusionism and eliminativism (Dennett, 1991; Frankish, 2016) attempt to dissolve the hard problem by claiming that PC is a cognitive illusion. Yet, as critics have noted (Strawson, 2018; Chalmers, 2020), these views still bear the burden of explaining why certain neural structures generate the illusion while others do not — effectively restating the attribution problem.

Property dualism and epiphenomenalism accept the existence of phenomenal properties but deny their causal efficacy, and they too must specify the conditions under which such properties are instantiated. Interactionist dualism would additionally require a causal criterion identifying where mental events exert non-physical influence on the physical world.

Even panpsychism, which treats consciousness as fundamental and ubiquitous, cannot fully escape discriminative work: it must solve the combination problem, namely, how micro-level proto-consciousness combines into macro-level unified subjects (Goff, 2019).

Thus, with the partial exception of the most radical panpsychist formulations ("everything is conscious"), virtually all differentiating theories ultimately require operational criteria. Without such criteria, they cannot be tested or applied to novel systems such as artificial agents.

## 2.2 Existing Discriminative Approaches

Several major attempts have been made to articulate discriminative frameworks:

**Integrated Information Theory** (IIT) provides a quantitative measure ($\Phi$) of informational integration and equates high-$\Phi$ states with consciousness. This makes IIT both a discriminative theory. However, its criterion is theory-laden: the identity claim "consciousness = $\Phi$" risks circularity, as it is not an independent empirical test (Cerullo, 2015; Bayne, 2018).

**The CHIP Test** (Schneider, 2019) proposes identifying the "minimally sufficient neural substrate" of consciousness by gradually replacing biological neurons with functionally equivalent components. This is powerful for mapping necessary conditions in biological systems but does not generalize to non-biological substrates.

**The ACT Test** (Schneider, 2019) is a more direct sufficiency test. It uses tasks requiring mastery of first-person phenomenal concepts (e.g., dreaming, inverted spectra) to probe whether a system truly possesses such concepts. ACT advances beyond Turing-style behavioral mimicry, yet its questions are still indirect: in principle, a sufficiently sophisticated non-conscious system might answer them correctly by statistical inference.

Taken together, these approaches highlight the promise and limitations of current discriminative work. They underscore the need for a substrate-independent, logically rigorous, and counterfeit-resistant sufficiency criterion.

## 2.3 Mainstream Neuroscience and the Ground-Truth Problem

Most neuroscientific and cognitive theories, including GWT, HOT, LRT, AST rely on neural-behavioral co-occurrence as evidence. For example, GWT infers that global broadcasting underlies consciousness because it co-occurs with reports of awareness (Dehaene, 2014). HOT theories depend on neuroimaging studies showing prefrontal activation during conscious report.

This methodology faces two core challenges:

**Correlation ≠ Causation**: Co-occurrence does not prove that the mechanism is sufficient for PC.

**Ground-Truth Circularity**: Using subjective report as the proxy for consciousness risks conflating "reportability" with phenomenality. No-report paradigms mitigate this but still require some criterion to classify trials as conscious or unconscious.

Thus, even empirically rich explanatory models remain hostage to the attribution problem: without a theory-neutral criterion for PC, they cannot validate their own success conditions.

## 2.4 The Urgency in AI Contexts

The rise of large language models (LLMs), multimodal agents, and world-model-based cognitive architectures has reignited debate about AI consciousness. The central challenge, however, is how to determine whether a non-biological system is phenomenally conscious. Mogi (2024), for instance, has proposed the concept of "conscious supremacy" as a way to identify tasks that only conscious processing can efficiently accomplish — an step toward an objective criterion. In parallel, other recent work has independently proposed criteria or theoretical conditions for consciousness and applied them to contemporary AI systems. For example, tests of whether LLMs might satisfy global-workspace-like dynamics (Goldstein & Kirk-Giannini, 2024) and whether functional architectures can instantiate integrated information (Albantakis et al., 2023). Empirical studies have begun to quantify which features lead humans to attribute consciousness to AI systems, for example, metacognitive self-reflection, emotional expression, and first-person perspective-taking (Immertreu et al., 2025, Li et al., 2025) and have also sought to formalize when generative models exhibit emergent "self-identity" (Lee, 2024). Hoyle (2024) integrates multiple existing theories (functionalism, IIT, and active inference) to analyze the internal mechanisms of OpenAI-o1, illustrating the prevalence of theory-dependent evaluation strategies.

These developments expose two urgent challenges for discriminative work:

**Behavioral Mimicry**: Advanced AI can convincingly claim to "feel" or "be aware" without providing any warrant that these claims track genuine phenomenality. Behavioral studies (e.g. Immertreu et al., 2025) reveal which cues humans rely on when attributing consciousness, but such attributions are vulnerable to counterfeiting by purely statistical models.

**Theory-Dependence**: Many proposals import specific consciousness theories—e.g., GWT (Goldstein & Kirk-Giannini, 2024), IIT (Albantakis et al., 2023), functionalist criteria for self-identity (Lee, 2024), and hybrid models (Hoyle, 2024)—to adjudicate AI consciousness. These approaches are informative but theory-laden: their verdicts depend on the correctness of the adopted theory, and thus cannot serve as a neutral arbiter.

The absence of a generally accepted, operational criterion has made mainstream functionalism a natural target of criticism. When functionalists assert that an artificial system performing the right kind of information processing should count as conscious, non-functionalists respond by asking: on what grounds? What independent criterion justifies such an attribution? This demand for justification has motivated a series of biological-grounding arguments: Seth (2025) cautions that phenomenal consciousness cannot be inferred from functional similarity alone and calls for a theory that accounts for its biological basis; Saad (2024) searches for a "biological crux" that would separate genuinely conscious systems from mere functional simulacra; Kleiner and Ludwig (2024) proposes a "no-go theorem" suggesting that consciousness cannot arise from non-neural substrates. These challenges highlight not the failure of functionalism per se, but the need for a criterion that can adjudicate such disputes in a non-stipulative, empirically testable way.

The lack of a criterion that can be applied across both biological and artificial systems risks reducing the debate to a purely stipulative one, rendering it empirically irresolvable. As Schneider (2019) emphasizes, what is currently missing is a theory-neutral, operational, and counterfeit-resistant test capable of empirically adjudicating such disputes. The persistent controversy between functionalism and anti-functionalism indicates that existing criteria are inadequate. This situation motivates the present study: to develop a sufficiency criterion that is both substrate-independent and logically rigorous, thereby transforming this philosophical disagreement into a scientifically testable question.

## 3. The simple interpretation of phenomenal consciousness

In this section, we will discuss the following:

1) A criterion for determining whether a machine is conscious from a third-party perspective.

2) Four principles of information processing based on Kant's idea.

3) A demonstration that a machine operating according to these principles meets the above criterion.

## *3*.1 How to judge whether a system has consciousness

The first challenge in explaining the nature of consciousness is determining how to judge whether a system is conscious from a third-party perspective. Without such a criterion, any explanation of consciousness would clearly be invalid, as it inevitably faces the question: why is the mechanism identified by this explanation equivalent to consciousness?

However, this first question is often considered extremely challenging. This is because one of the key characteristics of consciousness is its complete subjectivity—third parties cannot directly observe the consciousness of other subjects. In other words, consciousness is a subject's private information. But does being private information mean it is entirely unknowable?

In fact, people often have ways to judge whether another subject knows certain private information. One of the most basic methods is to provide part of the information to the subject and then ask them to supply the remaining part. For example, to determine whether a subject knows about a specific commemorative coin, the evaluator can describe one side of the coin and then ask the subject to describe the other side.

So, does this simple method apply to determining whether another subject possesses consciousness? To address this question, we need to further analyze what exactly we mean by "consciousness" in this context. Taking the *hard problem of consciousness* as an example, it asks: *How does a physical mechanism generate qualia?* However, this question actually encompasses two distinct levels of inquiry:

1) How a physical mechanism generates a specific qualia, e.g the feeling of redness (instance problem)

2) How a physical mechanism generates the category of qualia (category problem)

The *instance problem* and the *category problem*, while closely related, do not share exactly the same implications. For example, solving the *instance problem* could help answer questions like, "Is another person's feeling of redness the same as my feeling of redness?" On the other hand, solving the *category problem* might not provide an answer to this question, as it only determines what kind of system has the qualia but does not guarantee that two systems will generate the same qualia. Because these two problems have different implications, it is essential to distinguish between *qualia as category* and *qualia as instance*.

Now, we can return to our main topic—whether the usual methods for judging private information can be used to determine if another subject has phenomenal consciousness.

We argue that if phenomenal consciousness here refers to qualia as instance, then it cannot be determined. There are two main reasons for this. First, the description of a specific qualia always depends on other bodily sensations or actions, making it impossible to provide an objective physical description of the qualia itself. For example, describing the sensation of pain as the feeling of being pricked by a needle effectively attributes one sensation to another. This does not guarantee that the sensations described by different subjects are the same; it only ensures that there is a similar association between the sensations. Second, such descriptions inherently rely on the physical structure of the subject's body. If the bodily structures differ, can we still assume that the described qualia are the same? Therefore, we cannot use descriptive characteristics to determine whether two different subjects experience the same qualia.

However, the method of using characteristic information to judge whether a subject possesses certain private information can, in ideal circumstances, be applied to determine whether a subject possesses *qualia as a category*. This is because *qualia as a category* has many characteristics that can be objectively described and do not rely on a specific bodily structure. For example, consciousness is characterized by its ineffability, intentionality, and unity. Even the fact that *qualia as instance* cannot be reduced to an objective explanation is also an objective description of *qualia as a category* (physical irreducibility).

This also means that if a subject can describe key features of consciousness without having obtained information about consciousness from external sources, then it can be inferred that the subject possesses consciousness. (For example, large language models like GPT-4.0 would not be included in this category because their training data already contains a vast amount of information about consciousness.)

Some might question whether this method of general information assessment is appropriate for judging whether a machine has consciousness. They might point out that if we are to use the method described above to prove a machine has consciousness, what we need to verify is whether the machine can express multiple properties of consciousness. But it is possible that the machine does not truly have consciousness, but instead generates a concept internally that just happens to have the same properties as the verified properties of consciousness. In this case, the reason that the machine can express these properties would simply be a coincidence.

If we are to use the method described above to judge the credibility of whether a machine has consciousness to the same degree as we affirm our own consciousness, then we acknowledge that there are indeed flaws as pointed out. However, if we consider further, what reason do we have to judge that other people have consciousness? The best we can do is apply the method described above. This means that if a machine can express the same number of properties of consciousness as a human, we should believe that the machine has consciousness to the same degree of credibility as we believe other people have consciousness.

Therefore, we can conclude that a sufficient condition for determining, from a third-party perspective, whether a system possesses phenomenal consciousness is as follows: **If a system, without obtaining any information about consciousness from external sources, is still able to provide information about the key features of phenomenal consciousness as many as humans, we can determine that the system is conscious as confident as that we believe other persons have consciousness.**

It is worth noting that if this method of assessing consciousness is reasonable, then it implies the denial of the existence of philosophical zombies. This is because if a machine is functionally identical to a human, we should consider that it has the same consciousness as a human.

Moreover, it also denies the method of determining the existence of consciousness based on functional necessity. Specifically, if a machine, such as a speaker, has the ability to express the same number of consciousness properties as a human, someone might argue from a functional necessity perspective that the speaker can accomplish these properties mechanically and does not require consciousness. Therefore, they might claim it cannot be proven to be conscious. We believe such an argument is unreasonable, since its underlying assumption here is that consciousness is a functional component of humans. In fact, we should remain open to this assumption rather than treating it as a granted axiom.

## 3.2 Cognitive Principles of Information Processing: Inspired by Kant

Over 200 years ago, Kant presented his famous claim about human cognition in his *Critique of Pure Reason*. He argued that humans cannot directly access or understand the essence of things, but can only comprehend the surface of external objects through the sensations generated by their sensory organs. The sensations received by humans are then organized by their inherent, a priori structures. Ultimately, these organized sensations are imposed back onto the external world, thereby defining it.

In fact, if we set aside the philosophical implications, Kant's view is consistent with the basic understanding of humans in modern neuroscience: the human body forms perception by organizing the signals received by the sensory nervous system from the external world in an innately biased and experience-shaped way..

Taking this view as our starting point, we use the analogy of a machine enclosed in a room to represent the entire human neural system, including both the systems that receive signals from the external world and those that generate bodily actions. Moreover, it has the following characteristics:

1) On this machine, there is an array of signal lights, and all external information can only be transmitted into the room through these signal lights. However, the machine itself does not know how the sensors that transmit this information work, so it cannot deduce the true external reality based on illuminated lights.

2) The machine also has an array of buttons. At any given moment, under a specific state of the signal lights, pressing different buttons may lead to different patterns of signal lights illuminating in the future.

3) Additionally, the machine recognizes that some signal lights are special. Among these, some lights turning on are beneficial to the machine itself, while others are harmful. Therefore, the machine naturally seeks to maximize the illumination of beneficial lights while minimizing the illumination of harmful ones.

In this analogy, the signal lights correspond to the sensory signals humans receive. The buttons represent all the actions a person can perform, including body actions as well as non-physical ones, such as shifting attention. The special signals can be understood as events encoded in human genes that are beneficial or detrimental to the survival of the organism.

Imagine, in this scenario, how should the machine process the signals it receives to effectively increase beneficial signals while reducing harmful ones? One possible solution is to identify the relationship between past signals, button presses, and future outcoming signals, and then record this information. For example, the machine could learn rules such as:

*"If signal light 1 is on and button A is pressed, then signal light 2 will turn on next."*

If signal light 2 is beneficial to the machine, then whenever signal light 1 turns on, the machine effectively gains a way to make signal light 2 turn on, thereby increasing the likelihood of beneficial signals appearing.

However, due to the complexity of the external world, the likelihood that a single signal light can perfectly predict the activation of another single signal light is quite low. This would require two causally related external objects to correspond exactly to two specific signal lights. Clearly, this is not the effective way to organize information.

Compared to the above approach, a more effective way to accommodate predictive information is to use a group of past signals to predict another group of signals. Since a signal group can consist of just a single signal, this method not only includes cases where a single signal predicts another single signal but also allows for combinations of multiple signals predicting another group of signals. So, it is able to represent more rules than just the approach in which only one signal light predicts another

Following this approach, because both the predictive and predicted signal groups can exist independently of other noise signals, the causes in the external world that generate these signal groups can be considered as separate objects, isolated from their surrounding environment. This process of distinguishing objects from their context aligns with Kant's idea that the internal structure defines the external world.

Let's take Figure 1 as an example. In Figure 1a, the machine identifies the mutual predictive relationship between the visual and tactile signals of a football, allowing it to define the external concept of a football. In Figure 1b, by discovering the predictive relationship between rectangles and right angles in visual signals, the machine can define both right angles and rectangles.

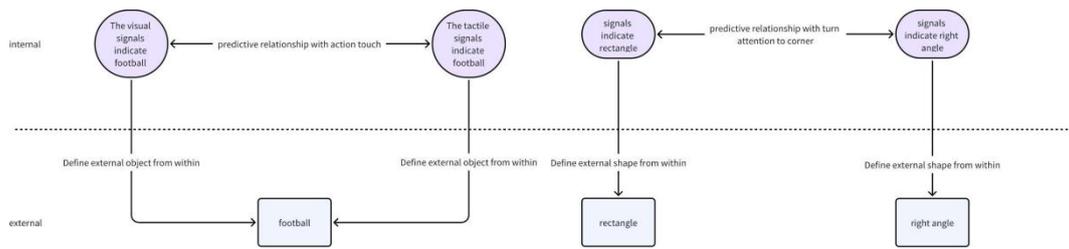

*Figure 1a: Forming the concept of football based on internal signals*

*Figure 1b: Forming the concepts of rectangle and right angle based on internal signals*

From this, we can derive the first principle of how a machine process information (**Prediction Principle**): The machine must identify and remember rules in which a combination of a set of signals and specific actions can predict the occurrence of another set of signals. Once these rules are discovered, the external causes that generate these two sets of signals will be defined as specific objects.

Next, we can further refine the concept of a "set of signals" mentioned in the above principle. In fact, when expanding from a single signal to a set of signals, there are two different dimensions of expansion. First, in the spatial dimension, a single signal can be extended into a collection of simultaneously occurring signals. Second, in the temporal dimension, signals from multiple past time points, along with specific actions, can collectively serve as the conditions or results for prediction.

We have already explained the reason for expansion in the spatial dimension—external objects with predictive relationships do not necessarily correspond to a single specific signal, so multiple signals can be combined to represent an external object. Now, let us further consider why temporal expansion is necessary. The reasoning is similar: signals occurring at a single moment may not sufficiently represent an external object with predictive relationships. Certain actions may further reveal information about the object. Therefore, the machine must extend sets of signals over time to better represent external objects.

We can summarize this temporal expansion as the second principle of information processing (**Exploration Principle**): When a combination of signals at a single time point is insufficient to establish an effective predictive relationship, the machine may explore additional information about the external object through specific actions, thereby forming a more effective predictive relationship.

Next, we consider the third principle. If a machine needs to live in a complex environment like a human, it should not treat all signals equally. This is because signals that are beneficial or harmful to it are more important than other signals. Therefore, it should prioritize processing signals related to these signals. We can consider such an example: if signal 1 is an inherently beneficial signal, and past experience shows that signal 2 can lead to signal 1 through press button A. Then, although signal 2 is not an inherently beneficial signal, the machine also has a way to obtain signal 1 through signal 2. Therefore, signal 2 becomes a signal related to the beneficial signal. So, in this case, the machine should also prioritize discovering other signals that can lead to signal 2. Through this chain-like organization, a signal network centered on inherently beneficial or harmful signals can be obtained, and this network gradually expands based on experience. The larger the coverage of this network, the more beneficial it is for the machine's survival.

Therefore, in addition to the previous two principles for organizing experiential information, we can also add the third principle **(Priority Principle)**: The machine should build predictive relationships that are related to its own survival with high priority.

With the principles outlined above, there remains a challenge: if the experiences summarized from the past are incorrect, those erroneous predictive rules need to be corrected. However, the prerequisite for making such corrections is the ability to pinpoint the specific flawed rules. Let's take an example to illustrate this. Suppose a machine detects signal 2, and based on past predictive rules, it has learned that pressing button 1 should yield a beneficial signal—signal 1. Yet, after pressing button 1, the machine finds that signal 1 does not appear. In this case, the machine needs to identify the rule that no longer applies.

However, if the machine cannot observe the internal process of how its signals are generated, even if it knows it has a rule stating "signal 2 plus button 1 leads to signal 1," it cannot locate this erroneous rule. This is because there might be multiple rules predicting that pressing button 1 will produce signal 1. For instance, another rule might state that "signal 3 plus button 1 also leads to signal 1." In such a situation, the machine cannot discern which specific rule led it to press button 1.

Therefore, the prerequisite for the machine to be able to locate and analyze erroneous rules is the ability to some extent to trace back to past signals. And this tracing back is not to trace back to all signals at a past point in time, but to trace back to a set of signals at a certain point in time (or time period) that have predictive power. In addition to this set of signals, it also needs some additional status signal lights to indicate that these signals were obtained through backtracking, rather than from the external stimulus.

From this, we can formulate the fourth principle **(Recall Principle)**: The machine has buttons that can reactivate previously occurring sets of signals that have predictive power, accompanied by a state signal indicating that these signals originates from itself.

From this, we have established the four information processing principles that a machine, constructed based on Kant's perspective, must follow to operate effectively based on past experience: the **Prediction Principle, Exploration Principle, Priority Principle,** and **Recall Principle**.

It is important to note that we are not claiming the four principles above exhaust all the principles by which a machine like a human processing information. They are merely the principles directly relevant to our discussion. There must be other principles for organizing experiential information.

For example, the Exception Recording Principle — which requires the machine to record exceptions to its predictions — is also desired. A machine must store information such as: If Signal 1 occurs and Action A is performed, Signal 3 can be predicted; however, if Signal 2 occurs simultaneously with Signal 1, then Signal 3 no longer follows.

However, principles like this are not directly related to the problem of consciousness that we aim to address next, so we will set them aside for now.

## 3.3 Why Is This Machine Phenomenally Conscious?

According to the method we proposed in the first subsection for determining whether a machine has consciousness, solving this problem requires identifying which object defined by the machine corresponds to human phenomenal consciousness or qualia. Furthermore, we must explain why this object possesses the same properties as human consciousness.

Therefore, let us first examine what kind of object should correspond to qualia. We can start by considering the following special predictive relationship. As stated in the Recall Principle, a specific button on the machine can reactivate past signals while attaching an indicator signal to show that these signals were retrieved through recall. In this case, as long as the past signals remain unchanged, the recalled signals will also remain unchanged. According to the Prediction Principle, this is a predictive relationship. Thus, the machine will define an object based on the signals obtained through recall.

We argue that the object defined in reverse by the signal set obtained through recall is qualia. To prove this, we need to verify that this object indeed possesses the core properties of phenomenal consciousness. Specifically, we will now examine its ineffability, physical irreducibility, intentionality, and unity one by one.

### 3.3.1 Ineffability and physical irreducibility

The ineffability of qualia refers to the idea that the subjective experience of external objects cannot be objectively described. To understand why the object defined by recall cannot be objectively described, we first need to understand what objective description is.

Let's take Figure 1b as an example to explain what objective description means. The relationship between right angles and rectangles depicted in Figure 1b is widely accepted as an objective description (A rectangle has four right angles). The key characteristic of such a relationship is that it remains consistent across different machines, as long as they can represent the shapes of rectangles and right angles, even if their internal representations differ.

The reason for this is that, although different machines may have different internal signal representations for a rectangle and a right angle, when they learn the terms "rectangle" and "right angle," they have already bound these terms to their respective internal signals. Therefore, even though the signal sets triggered by these machines when observing the relationship between rectangles and right angles may be different, their expression of the relationship will be the same. In other words, the core of this consensus is the signal alignment process that occurs when defining external objects in language. *Thus, the so-called objective description is actually a description of the relationships between objects that can make signal alignment with each other.*

Let's reconsider the object defined through the recall action. Clearly, this object does not have an external object as its trigger because it is actually triggered by past signals within the system. These signals are internal, so they cannot be aligned according to external objects (as shown in figure 2). This explains why the objects defined through the recall action cannot be objectively described.

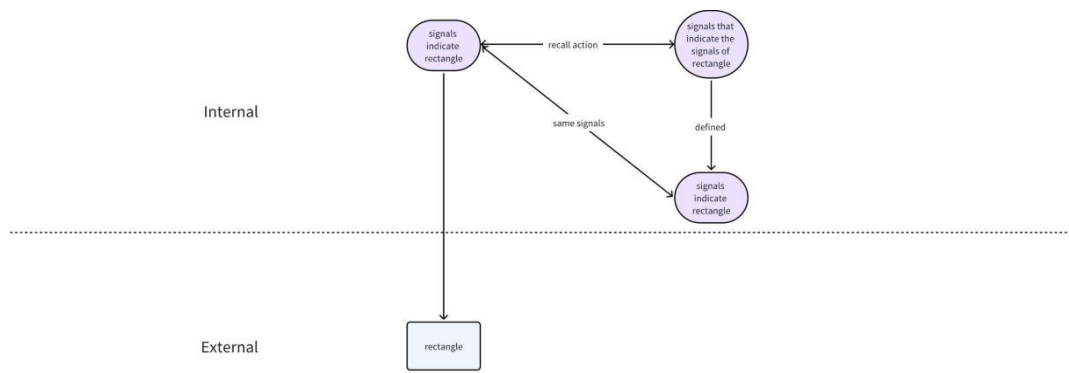

Figure 2: Forming the concept of qualia of rectangle

Furthermore, since the object defined through the recall action cannot be objectively described, it must therefore be physically irreducible. This is because the physical reducibility is explaining one objective phenomenon through another objective phenomenon (all physical phenomena must rely on objective language as a foundation). The inability to be objectively described implies that the phenomenon cannot be explained in terms of other physical phenomena. Therefore, it must be physically irreducible.

### 3.3.2 Intentionality

The definition of the intentionality of consciousness is that human phenomenal consciousness is always about a specific thing. In fact, this is naturally true for the machine we are discussing. As we know from the recall principle, the object defined by recall is a set of signals with predictive power. Additionally, as we know from the prediction principle, a set of signals with predictive power is used to define external objects in reverse. Therefore, we can conclude that the object defined by recall (the object we argue corresponds to human qualia) inevitably indicates a specific external object. Thus, the object defined by recall has intentionality.

Representationalists might object that our treatment of intentionality is overly simple, since intentionality is often taken to involve not only directedness but also accuracy conditions (sometimes expressed as being truth-evaluable). For example, when one sees a red apple, the content of the experience can be expressed as "there is a red apple there." If there really is a red apple, the experience is accurate; if it is in fact a red lamp, the experience is illusory.

At first glance, this requirement might seem to go beyond the principle we have formulated. Yet, once we ask how we ever confirm that an apple is "really there," the connection becomes clear: we do so by further testing—touching it, viewing it from different angles, or checking again later. All these procedures amount to verifying the validity of the predictive relationships linking current perception to past and expected sensory patterns. When these predictions are upheld, we treat the experience as accurate; when they systematically fail, we treat it as inaccurate. Thus, intentionality's capacity for truth/falsehood is fully anticipated by our principle.

On this view, "truth" is not defined as correspondence with an unknowable noumenal world, but as consistency with the subject's broader web of experience and prediction—a perspective that resonates with Kant's insight that the world-for-us is constituted through our forms of cognition.

### 3.3.3 Unity

In this paper, by "unity" we adopt the standard notion of phenomenal unity (Bayne, 2010), namely, the integration of multiple concurrent experiences into a single, coherent phenomenal field. We acknowledge that some philosophers discuss broader notions of unity, such as diachronic unity or self-related unity, but these are beyond the scope of our present criterion, which focuses on the synchronic structure of phenomenal consciousness.

Does this apply to the object defined by recall? While it's true that at each moment, all the signals in the machine may not necessarily be grouped into a single object (meaning they aren't treated as a whole), when the machine is asked to reflect on its own experience, it must employ the recall principle. The objects recalled by the recall principle must be sets of signals with predictive power, and these signal sets are used to define external objects. As a result, the machine will inevitably perceive its past experience of consciousness as a unified whole.

However, the fact that a machine perceives signals as a unified whole does not ensure that it recognizes this perception as a unified experience. For the machine to understand its experience as a unified whole, it must be aware of both the whole and its individual components. Once it grasps this, it can identify the relationship between them, leading to the conclusion that the overall experience is composed of multiple components.

Specifically, if a machine has the ability to activate only a single source of sensor at a time, it can isolate individual components. For example, a machine equipped with both auditory and visual perception can activate only its auditory sensors to capture the auditory component or only its visual sensors to capture the visual component. If the machine also possesses the basic ability to recognize when one set of signals is a subset of another, then it can naturally determine that a group of signals involving both hearing and vision consists of auditory and visual components. As a result, this enables the machine to recognize that its consciousness（group of signals）is a unified whole composed of signals from various types of sensors.

### 3.3.4 From Knowing to Speaking

Now, we have explained that a machine operating based on the four principles would define and know a category of objects used to represent a set of signals. These objects have the same four key properties as human phenomenal consciousness: ineffability, physical irreducibility, intentionality, and unity. However, there is still a potential issue: according to the criteria we previously established for determining whether a machine is conscious, a machine needs to be able to articulate these properties, not just know them. So, is there an obstacle in moving from knowing to expressing these properties?

We believe that there is no such obstacle. This is because the principle by which the machine defines an object with the same properties as qualia is entirely consistent with how it defines other ordinary objects. Therefore, as long as the machine is capable of perceiving and expressing the properties of ordinary objects, it should also have the ability to express the properties of this special object. However, it is clear that the machine's ability to express the properties of ordinary objects is unrelated to whether it has phenomenal consciousness. It is instead related to the machine's capacity for detecting phenomena and its expressive ability. In other words, this is an issue of intelligence, not of whether consciousness exists. Thus, we believe that as long as the machine has sufficient intelligence and expressive ability, it will have no problem expressing the properties of an object corresponding to human phenomenal consciousness.

Based on the above discussion and criteria, we can conclude that a machine operating according to the four principles possesses phenomenal consciousness. Moreover, the degree of credibility is consistent with our belief that other humans possess consciousness.

# 4. Bridging the Gap Between Philosophical Principles and Empirical Evidence

The principles of the *conscious machine* we introduced earlier is derived through rational reasoning. A natural question then arises: do humans operate based on the same principles? We argue that the answer is affirmative. However, establishing this claim is far from straightforward. The difficulty lies in the fact that our methodological approach differs from that of mainstream cognitive science. Specifically, our reasoning proceeds in a top-down manner, aiming to construct a conscious intelligent machine by deducing the necessary conditions for its realization. In contrast, mainstream cognitive science primarily seeks to explain isolated phenomena observed in human cognition. If empirical evidence from cognitive science is to be used to assess whether the proposed principles also apply to humans, it is first necessary to establish a correspondence between the terminologies employed within these two distinct methodological frameworks.

## 4.1 Prediction Principle and Binding features

In cognitive science, binding features refers to the process of integrating sensory information or features into a unified perceptual object. This concept bears similarities to the group signals in the prediction principle. Both describe the process of binding externally processed sensory signals together to be perceived as a whole. However, there are also differences between the two. In mainstream theories, the cause of feature binding is primarily attributed to attention. Features that are attended to simultaneously are bound together. In contrast, according to the prediction principle, grouping signals arise because this set of signals has predictive power. These signals can anticipate the appearance of specific signals, especially those related to phenomena that are of personal relevance or interest.

So, is there a relationship between the two? The answer is yes. Specifically, grouping signals based on predictive power serves as the goal, while binding features based on attention functions as the mechanism to achieve this goal. Binding features essentially capture the characteristics of features occurring together, which forms the basis for group signals. In comparison to grouping signals, binding features lacks the aspect of whether the group of bound features has sufficient predictive power. Therefore, a plausible hypothesis is that the brain first binds a large number of features and then, based on the predictive ability of these bound features—especially their predictive relevance to the individual—retains those features that can make important predictions and their associated predictive relationships, while discarding those bindings that are not important.

Although there is currently no direct evidence confirming that binding features is indeed a step in the process outlined above, some evidence provides preliminary support for this idea. For example, there is evidence suggesting that the strength of binding features does indeed vary depending on whether it can predict important events related to the self. Specifically, features that are directly related to a task or that affect emotions—emotions are often seen as signals of events that are either harmful or beneficial to oneself (Lazarus, 1991; Lerner & Keltner, 2000; )—tend to have their bindings enhanced (Phelps, 2016; Barbot & Carrasco, 2018). This enhancement is achieved through the allocation of more attention or more frequent occurrences of these features (Yiend, 2009; Mattson, Fournier, Behmer, 2012). The phenomenon of enhancing feature bindings relevant to the individual aligns with the predictive network posited by the "prediction principle" and the "priority principle," which suggest that the brain, through experience, forms signal groups associated with the individual's interests and concerns.

## 4.2 Exploration principle and Attention

Once we clarify the relationship between binding features and grouping signals based on predictive power, we can further compare the process of binding (see Figure 3).

The key difference between the two processes lies in their perspectives on attention. In the mainstream view, attention itself is considered a mechanism that enables different features to be bound together. For example, the Feature Integration Theory (FIT) proposed by Treisman and Gelade (1980) states that the visual system can only access simple feature maps—such as color, direction, and motion—during the preattentive stage. These features are formed through bottom-up processing. As a result, detecting a salient target defined by a single basic feature, such as a unique color or direction, is relatively easy. However, tasks involving feature binding require selective attention, as binding multiple features, perceiving details, and most object recognition tasks all depend on this mechanism.

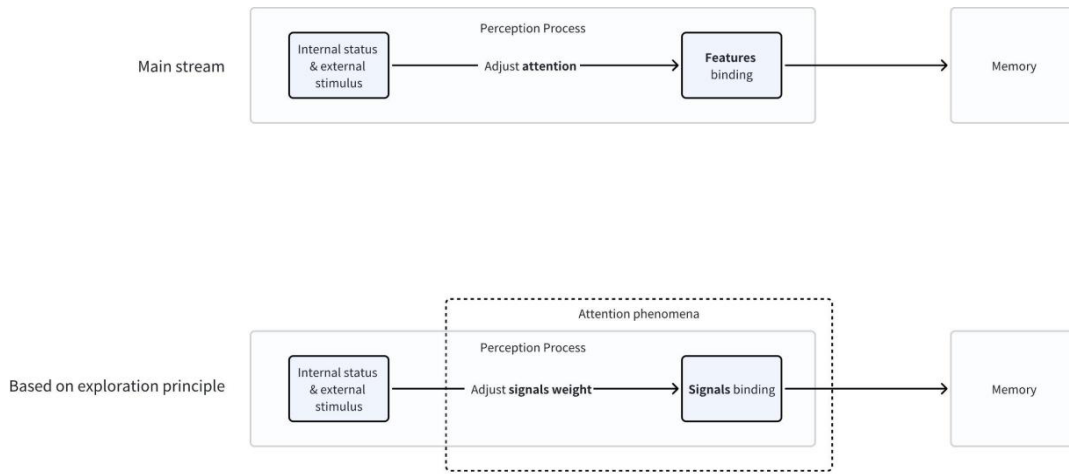

Figure 3: The main stream's idea of perception process and the perception process based on exploration principle

However, in the mechanism based on the exploration principle, there is no dedicated mechanism equivalent to attention. Instead, the system adjusts the weights of signals, allowing certain signals to be displayed while others are suppressed. Information that is simultaneously displayed naturally undergoes signal binding.

Taking our previously discussed simple machine as an example: given a specific external condition, the array of signal lights in the machine is in a certain state. Pressing a button activates new signal lights, which is analogous to acquiring new information. These simultaneously illuminated signals begin to undergo an initial binding process on their own. This binding serves as a preparatory step for the later selection of bindings with predictive power.

We can see that the attention mechanism and the process of changing signal lights through button presses share certain similarities. In terms of purpose, both serve to help the agent acquire more information. In terms of implementation, both can involve receiving signals caused by alterations in the sensory apparatus—such as moving the body or rolling the eyeballs—and can also include purely neural processes, such as top-down neural control, which amplifies certain signals while suppressing others, without requiring direct bodily involvement.

Although these similarities exist, there are three key differences between them. We will now examine each of these differences in detail.

First, in mainstream theories, attention is typically viewed as focused. This implies that feature binding only occurs when attention is directed toward specific features.

In contrast, a machine based on the exploration principle does not distinguish between a focused and a dispersed state—as long as signals are activated simultaneously, binding begins to occur.

The idea that feature binding can occur without the need for focused attention is supported by some recent empirical findings (Balas, Nakano & Rosenholtz, 2009; Rosenholtz, 2024; Keshvari & Rosenholtz, 2016).

Second, in mainstream Feature Integration Theory, attention binds only simple features such as color, direction, and motion. In contrast, the conscious machine does not impose such restrictions — any signal represented by a signal light can be a binding candidate.

In fact, according to the four principles we proposed, the machine itself lacks the ability to distinguish between simple and complex features. It merely receives signals and organizes them in a fixed pattern. Empirical studies have shown that binding in cognition is not necessarily limited to simple features — some bindings involve more complex and meaningful objects (Joseph & Nakayama, 1997; Rosenholtz et al., 2012; Cohen, Alvarez, & Nakayama, 2011; Ehinger & Rosenholtz, 2016).

In fact, considering the previous two points, we argue that the mainstream dichotomy between a preattentive stage and an attentive stage is questionable. Instead, we propose that the process of information exploration in perception is better understood as mode switching.

Taking our simple machine as an example, when observing the external environment, the machine initially operates in a mode where no button has been pressed — this can be referred to as the default mode. When the information acquired in this mode proves insufficient for completing a task, the machine switches to another mode for further exploration.

Human cognition follows a similar manner. What mainstream theories describe as the preattentive stage corresponds to the default mode, while attention corresponds to a switch to other focused observation modes.

Rosenholtz's (2024) dual-task experiment provides strong support for this view. In the experiment, even very simple tasks were found to interfere with each other's performance. According to the preattentive and attention stage dichotomy, these tasks should have been processed in parallel and thus should not have affected each other. However, under the mode-switching hypothesis, since the two tasks require different observational modes, they inevitably interfere with one another.

If the mode-switching theory is correct, then the traditional concept of an attention mechanism may not exist as an independent cognitive function. Instead, what we call attention is merely a phenomenon that occurs when the switched mode happens to be a focused observation mode.

Finally, and most importantly, regardless of whether the perceptual process relies on an attention mechanism or mode switching, there remains the fundamental question of how to explore effectively. In other words, why is attention almost always directed to the most relevant locations? The standard explanation is that attention allocation results from the coordination between bottom-up and top-down neural processes. However, this explanation is insufficient because it lacks a concrete mechanism to ensure the effectiveness of attention allocation.

On the other hand, if the mind possesses a predictive relationship network built upon grouped signals, it can act as a guide for how to allocate attention to specific states. This is because the grouped signals encompass both the relationships between external signals and the actions required to obtain further relevant information.

Let's illustrate this scenario with an example. For instance, in a search task where the participant is asked to find a "T" among a pile of "F"s, if the "T" happens to appear in a location that can be directly distinguished, it is equivalent to having a set of signal lights in the array indicating the presence of a "T" at a specific position. The subject can then directly identify that location based on this information.

If the "T" appears in a location that cannot be directly distinguished, the subject can rely on signals that represent the potential presence of a "T" (i.e., there is a predictive relationship associated with this signal suggesting further observation to confirm it as a "T," and another predictive relationship suggesting that, if further observed, it could be an "F"). These signals not only support the potential existence of a "T," but also include the mode-switching information on how to verify whether it is truly a "T" (similar to pressing the button in the previous example). Thus, the subject can use these cues to make further confirmations. This process of switching observation modes based on grouped signals is essentially the process of directing attention. In this process, the actions contained within the grouped signals provide guidance on how to explore further.

Although there is currently no direct evidence to support the above method of exploring information through grouped signals, one phenomenon derived from it has been widely supported by empirical evidence. That is, past experiences can alter the way individuals perceive things.

For example, Kundel and La Follette (1972) found that radiologists' visual search patterns when reading X-rays showed that experienced doctors could identify abnormal areas more quickly, reflecting the impact of professional experience on observation patterns. This means that task histories related to professional expertise deeply shape perceptual processing. Reingold and Sheridan (2011) reviewed the visual expertise skills of chess players and doctors, emphasizing how task experience shapes individuals' observation and perception. Werner and Thies (2000) compared the performance of football experts and novices in detecting changes in football scenes, finding that experts were more sensitive to key changes, supporting the influence of task experience on observation patterns.

The evidence for perceptual differences induced by historical tasks is not only reflected in past professional experience, but also in cultural or recreational activities. For example, Nisbett and Miyamoto (2005) reviewed how cultural backgrounds influence individuals' ways of perceiving, pointing out that East Asian cultures tend to emphasize holistic perception, while Western cultures are more inclined toward analytical perception, reflecting the impact of long-term tasks and environments on observational habits. Boot et al. (2008) found that individuals who frequently play video games exhibit higher attention and executive control in various cognitive tasks, suggesting that specific task experiences can alter observation and perception habits.

These pieces of evidence suggest that the brain must possess a mechanism that integrates past experiences with current modes of observation. Our previous discussed scenario provides a plausible method for solving this integration issue.

## 4.3 Recall Principle and Memory Indexing Theory

With this, we have demonstrated the relationship between the prediction principle, priority principle, exploration principle, and existing concepts in cognitive science. The recall principle has a more direct connection. The recall principle states that the objects of recall are a group of signals, which aligns with the hippocampal memory indexing theory. Memory representations are not stored in the hippocampus itself but are distributed across cortical areas, with the hippocampus serving as a pointer to these regions. When people recall something, these pointers are used to reactivate the features that were bound together in the past (Teyler & DiScenna, 1986; Teyler & Rudy, 2007). Based on the relationship between grouped signals and binding features discussed earlier, this suggests that, at least in long-term memory, it is consistent with the recall principle. Additionally, some recent studies have indicated that short-term memory also involves an indexing mechanism (Fiebig, Herman & Lansner, 2020).

## 5. Further Clarifications and Open Questions

At this point, we have presented a machine that satisfies the sufficiency criterion proposed in Section 3.1 and have provided a preliminary argument that humans are likely to be consciousness machines governed by similar principles. However, several important questions concerning consciousness remain to be addressed. In this section, we discuss some of these open questions and clarify the scope of our framework.

*Is a Conscious Machine Aware of All Events that Occur Within It?*

No. A system is conscious of an event only if the state of that event is recorded. If an event's signals are never recorded—or are immediately lost—then even if they momentarily activated a signal group capable of generating phenomenal consciousness, the system would have no memory of the event and thus no basis for reporting being conscious of it.

Blindsight provides a well-known example: blindsight subjects can navigate a maze flawlessly while insisting they see nothing. In our framework, this means the signal groups capable of producing consciousness were activated, but their activation sequences were never stored as objects in memory. More generally, we suggest that human subconscious activity could be driven by potentially conscious signal groups whose activations are not recorded or are quickly forgotten.

*Can Our Theory Distinguish Conscious from Non-Conscious Systems?*

Another question is whether our framework provides a principled distinction between conscious and non-conscious systems. Strictly speaking, it does not. As discussed in Section 3.1, the criterion we propose is a sufficient condition for phenomenal consciousness, not a necessary one. Consequently, the fact that a system fails to meet our criterion does not entail that it lacks consciousness.

Similarly, the framework and principles described in Section 3.2 offer only one possible way of satisfying the criterion. They do not exhaust the space of possible implementations. Other architectures or mechanisms, not captured by our framework, may also instantiate phenomenal consciousness. The only conclusion we can draw with certainty is that any system satisfying our four principles must possess phenomenal consciousness.

*How Can We Explain a Specific Subjective Experience, Such as the Feeling of "Greenness"?*

For phenomenal consciousness, one of the most troubling and most important questions for a long time has been: How to explain the subjective experience like the color green or joy as the result of a physical process? People generally hope for an answer in which a specific mechanism corresponds to a specific feeling. However, unlike what is commonly imagined, our answer to this question is that, in principle, the question is unsolvable. This view may seem difficult to accept: why can all other phenomena, except for the most basic physical phenomena, be explained, but phenomenal consciousness cannot? Yet, if we look at this from another perspective, it actually turns out to be a natural result.

Consider a machine constructed according to the principles we have described above. If it can faithfully record the sequence of state changes of its own past signals, and can re-evoke these states under certain conditions, then it is evident that this machine knows that certain internal states exist. However, it is unable to express what these states actually represent of the external world, because it does not know how these signals were generated.

According to our earlier assumptions, this machine can only represent an object by describing the relationships among these groups of signals. When even such descriptions cannot be made, then explanation becomes out of the question. (This is because explanation itself is the process of combining a set of experiential descriptions to infer another experiential description. For example, when we explain why a brick falling from a height shatters glass, we might say that the speed of the brick increases as it falls, and that a fast-moving hard object striking glass will shatter it. Here we used two experiential descriptions: (1) a brick falling from a height increases its speed, and (2) a fast-moving hard object striking glass will shatter it. By combining these two experiential descriptions, we explain the experiential description that a brick falling from a height shatters glass.)

In this context, the activation of a signal group essentially serves as an indicator of the system's internal state. This understanding of qualia is somewhat close to Donald Hoffman's(2019) view that qualia function as an interface: they do not represent any objective reality but serve as abstract symbols. When a system is capable of recording the past activations of these indicators and comparing them, it becomes able to distinguish that genuine state changes have occurred within itself. Furthermore, if the system is capable of categorizing these internal states as belonging to a common class, it will realize that instances of this class cannot be explained in terms of any other external objects—because they are precisely the foundation on which external objects are constructed. We simply give this class of states a name: "subjective experience."

In summary, if humans are machines that operate according to the principles we have discussed above, then human capacities for description and explanation must rest upon certain foundational elements. Since these elements cannot themselves be described in terms of other objects, they cannot be explained, predicted, or empirically verified.

# 6. Conclusion

In this paper, we presuppose such a machine:
1) It has no pre-defined concepts; it can only summarize patterns among external signals and organize these patterns into concepts.
2) Its ability to discover patterns is so powerful that it can establish the very category of "signals" itself.

Then, this machine will discover that the object representing the signals themselves cannot be interpreted by any other concepts. This conclusion has significant philosophical and practical implications simultaneously.

## 6.1 Philosophical implication

### 6.1.1 About the completeness of the physical world

From a philosophical standpoint, the difficulty humans face in explaining consciousness stems not from any incompleteness of the physical world (such as the need to invoke dualism), but from the inherent limits of explanatory capacity itself. Even if an intelligent agent exists in a purely physical world, its explanations of phenomena are constrained by the structure of its own thought processes, which operate on a set of basic representational elements. Explaining phenomena in physical terms is essentially a matter of establishing logical relations among objects composed of these elements. Consequently, it is neither surprising nor problematic that these basic elements cannot themselves be explained by the very objects constructed from them—this is a normal and inevitable outcome of the explanatory framework.

This view differs fundamentally from Colin McGinn's *mysterianism* (1989), which attributes the explanatory gap in consciousness to evolutionary contingencies: our cognitive architecture, as shaped by biological evolution, lacks the capacity to connect subjective experience with physical processes, though in principle a different evolutionary path might produce beings with such a capacity. In contrast, the present argument does not depend on evolutionary history. Once a cognitive system—human or otherwise—develops the capacity to represent the very signals it receives, it will inevitably encounter the same structural limit: the basic elements of its representational framework cannot be explained in terms of the objects constructed from them. This is a logical inevitability, not a contingent outcome of evolution, and it would apply to any cognitive architecture, regardless of its origin.

In this respect, the position advanced here is closer in spirit to Gödel's incompleteness results for formal systems than to evolutionary psychology. Just as a sufficiently expressive formal system cannot prove all truths expressible within it, a sufficiently capable representational system cannot fully account for the fundamental elements from which its own representations are constructed. This structural constraint offers a philosophically grounded explanation for why consciousness resists complete self-explanation, without requiring any departure from a purely physicalist worldview.

### 6.1.2 About the reality

Another important implication concerns whether we can regard qualia—these fundamentally inexplicable objects—as are mere illusions. In fact, this depends on how we define what is real. However, regardless of which definition of "reality" is adopted, it is logically contradictory to affirm that objects composed of fundamental elements are real while denying the reality of those very elements themselves.

In everyday discourse, the term "existence" typically carries two distinct meanings:
1）Perceivable objects: If an object can be perceived, then it exists. In other words, its reality is determined by whether it can be perceived.
2）Existence only if the object exists objectively: The existence of an object does not depend on whether it is perceived. In fact, this corresponds to Kant's notion of the thing-in-itself.

If the first definition is adopted, phenomenal consciousness serves as the basis of perception, and the existence of perceivable objects depends upon this foundation. If the second definition is adopted, sensation is merely a representational mode of external signals; however, such objective existence cannot be directly known, and thus denying phenomenal consciousness entails denying the determinacy of the existence of all objects.

## 6.2 Practical Implication

Setting aside the philosophical implications discussed above (that is, assuming the physical world really is as we know it), the principles proposed in this paper also address a question we consider very important: if the nervous system has a single unified purpose of integration, then what is that purpose? We believe that the answer to this question is extremely valuable for understanding human thought and the overall functioning of the nervous system. This is because the question is neither as trivial and complex as focusing on the details of various neural mechanisms, nor as overwhelming as analyzing the endless diversity of human behavior. Therefore, it is likely to become a breakthrough for understanding human cognition as a whole.

This point is actually quite similar to having a person who knows nothing about how computers work try to explore the principles behind them. Whether they start from the computer's inputs/outputs and its appearance, or the operation of semiconductor components, trying to uncover the true working principle of the computer is relatively difficult. However, if one begins with the concept of the Turing machine as a mathematical prototype, then grasping the principles of modern computers becomes relatively easy.

## Declaration of generative AI and AI-assisted technologies in the writing process

During the preparation of this work, the authors used Google Gemini, xAI's Grok, and OpenAI's ChatGPT to translate, check grammar, and polish the language for clarity. After using these tools, the authors reviewed and edited the content as needed and take full responsibility for the content of this article.